\pdfoutput=1

\documentclass[11pt]{article}

\usepackage[]{acl}

\usepackage{graphicx}
\usepackage{booktabs}
\usepackage{pgfplots}
\usepackage{url}

\usepackage[T1]{fontenc}

\usepackage[utf8]{inputenc}

\usepackage{microtype}

\title{Towards Domain-Independent Supervised Discourse Parsing Through Gradient Boosting}

\author{Patrick Huber and Giuseppe Carenini\\
  Department of Computer Science \\
  University of British Columbia \\
  Vancouver, BC, Canada, V6T 1Z4 \\
  {\tt \{huberpat, carenini\}@cs.ubc.ca}}
        
\begin{document}
\maketitle

\section{Introduction}
Discourse analysis and discourse parsing have shown great impact on many important problems in the field of Natural Language Processing (NLP) (e.g., \citet{ji-smith-2017-neural,bhatia2015better,nejat-etal-2017-exploring,gerani-etal-2014-abstractive}). Given the direct impact of discourse annotations on model performance and interpretability, robustly extracting discourse structures from arbitrary documents is a key task to further improve computational models in NLP. To this end, a variety of complementary discourse theories have been proposed in the past, such as the lexicalized discourse framework \cite{webber2003anaphora}, the Segmented Discourse Representation Theory (SDRT) \cite{asher1993reference, asher2003logics}, and the Rhetorical Structure Theory (RST) \cite{mann1988rhetorical}, with RST focusing on the semantic and pragmatic structure of complete monologue documents, as used in this work. 

\begin{figure*}
    \centering
    \includegraphics[width=\linewidth]{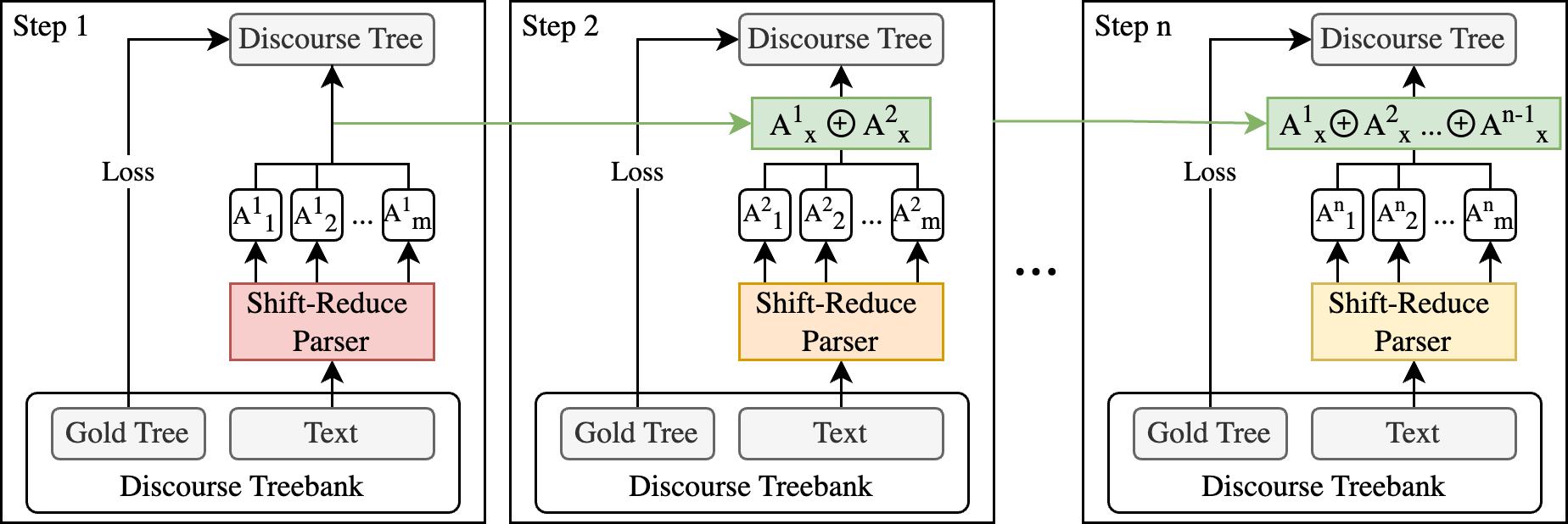}
    \caption{Schematic of our proposed gradient boosted discourse parsing model. Grey=Inputs/Outputs, Green=Gradient Boosting component, Red/Orange/Yellow=Trained models. A\textsuperscript{1}\textsubscript{2} refers to the second shift-reduce action of the first step. Please note that the models are independent and only connected by the aggregation in green.}
    \label{fig:grad_boost}
\end{figure*}

Despite the importance of discourse analysis and discourse parsing for the field of NLP and the obvious value of the RST discourse theory for many downstream applications,
one major limitation for a wider application of discourse information is the severe data sparsity issue (for instance, the popular RST-DT \cite{carlson2002rst} and GUM \cite{Zeldes2017} treebanks do not exceed a minuscule number of 400 documents). Furthermore, while the data sparsity issue has been a long-standing problem, modern, data-intensive machine learning approaches further reinforce its severity.

In general, three modelling alternatives have been established in the current landscape: (i) Supervised approaches (e.g., \citet{ji-eisenstein-2014-representation, feng2014linear, joty-etal-2015-codra, li2016discourse, wang-etal-2017-two, guz-etal-2020-unleashing}), performing well in the domain in which they are trained, however, obtain severely reduced performance if a domain shift is present, as shown in \citet{huber-carenini-2019-predicting, huber-carenini-2020-mega}. (ii) Distantly supervised models (e.g., \citet{huber-carenini-2019-predicting, huber-carenini-2020-mega, nishida-nakayama-2020-unsupervised, karimi-tang-2019-learning, liu-etal-2019-single, huber2021predicting, xiao-etal-2020-really}), aiming to overcome the domain adaptation problem by exploiting large-scale supervised datasets from context-sensitive auxiliary tasks (e.g., sentiment analysis). 
(iii) Self-supervised/unsupervised methods (e.g., \citet{zhu-etal-2020-examining, koto-etal-2021-discourse, wu-etal-2020-perturbed, kobayashi-etal-2019-split, huber2021unsupervised, huber-carenini-2022-towards}), predicting discourse from either pre-trained language models, auto-encoder style frameworks, or by recursively computing dissimilarity scores.

In this landscape of models aiming to overcome the data sparsity and domain dependency of current discourse parsers, 
we present a new, supervised paradigm directly tackling the domain adaptation issue. Specifically, we introduce the first fully supervised discourse parser designed to alleviate the domain dependency through a staged model of weak classifiers by introducing the gradient boosting framework \cite{NIPS1995_b4d168b4, DRUCKER199453, NIPS1997_9cb67ffb, badirli2020gradient} into the process of discourse parsing. Using the underlying assumption that any discourse treebank contains a mix of frequently appearing, general discourse features (applicable to any domain) as well as a number of dataset-related nuances (which are domain-specific), we postulate that a set of weak classifiers is likely to learn increasingly specific and rare features of the training data. Using this assumption, we can reasonably assume that there exists a threshold of weak classifiers, which effectively separates the general features of discourse from domain-specific characteristics introduced by the dataset. As a result, we aim to separate this mixture of features using the gradient boosting approach with the goal to generate a more domain-independent discourse parser.

\section{Approach}
\label{sec:approach}
Our approach to introducing neural gradient boosting into the domain of discourse parsing builds on top of the state-of-the-art (SOTA) neural shift-reduce parser by \citet{guz-etal-2020-unleashing, guz-carenini-2020-coreference}\footnote{For more information on the underlying approach, we refer interested readers to \citet{guz-etal-2020-unleashing}.}. 
In this work, we aim to extend this previous line of research in three meaningful directions:

\subsection{Added Discourse Relations}
The method proposed in \citet{guz-etal-2020-unleashing, guz-carenini-2020-coreference} reaches SOTA performance on the RST-DT structure and nuclearity prediction, however, does not consider the important relation attribute. In this work, we aim to generate complete discourse trees with all three components, introducing an additional relation-prediction component besides the structure and nuclearity predictor.

\subsection{Linguistically Inspired Stack Representations}
In the current SOTA method, spans on the stack are truncated by removing tokens from the center of the textual representation (e.g., a sequence of $\{t_1, t_2, t_3, t_4\}$ and a maximal length of $2$ results in $\{t_1, t_4\}$). As shown by the promising performance of the approach, this heuristic assumption seems reasonable, however, lacks linguistic justification. To this end, we propose a new method to reduce stack elements according to the sub-tree nuclearity. This directly follows the argument in \citet{morey-etal-2018-dependency}, stating that the relation between constituents in an RST-style discourse tree holds between the respective nuclei of the sub-trees.

\subsection{Gradient Boosting Approach}
Gradient boosting refers to a classical machine learning approach using an ensemble of weak classifiers initially developed for decision trees \cite{NIPS1995_b4d168b4} and later adopted for neural architectures \cite{NIPS1997_9cb67ffb, badirli2020gradient} (oftentimes called neural/deep gradient boosting), which has been shown to benefit important NLP tasks, such as part-of-speech tagging \cite{abney-etal-1999-boosting}, sentiment analysis \cite{athanasiou2017novel}, and text classification \cite{kudo-matsumoto-2004-boosting}, delivering robust models when data is scarce. Our model architecture envisioned in this work is presented in Figure~\ref{fig:grad_boost}. Following the gradient boosting paradigm, we start with a single weak classifier (left side in Figure~\ref{fig:grad_boost}) and train a standard shift-reduce model to predict RST-style discourse trees. The number of free parameters, purposely chosen to be small for individual, weak classifiers, is thereby likely to limit the ability of the model to learn complex features and relations, resulting in the initial training step to exploit simple structures, e.g., resembling purely right-branching trees. After convergence of the initial weak classifier, a second step is introduced with a similar-sized set of free parameters in the shift-reduce parsing component. This time (see the center in Figure~\ref{fig:grad_boost}), the parsing component is trained to improve the performance of the combined prediction consisting of the initial parser in step 1 and the currently trained component. With the parameter-frozen prediction from the first step being combined with the output of the parser in step 2, the combined model is bound to learn more nuanced relationships in the data. Following the gradient-boosting methodology, the second step thereby improves (i.e., boosts) the performance of the joint classifier for samples that the first step did not capture. Repeating this process for $n$ times, an increasingly specific parser is built. 

To summarize, we believe that the gradient boosted method in combination with our extensions of the SOTA work by \citet{guz-carenini-2020-coreference} should improve the domain-independence of supervised discourse parsers when trained on small-scale, human-annotated discourse treebanks. 
With the iterative modelling strategy, our gradient boosted method can likely utilize the limited training data more efficiently, achieving more domain-independent models, while still reaching high performance in-domain.

\section{Planned Evaluation}
Following our novel extensions proposed in section \ref{sec:approach}, we plan to evaluate the model along four dimensions: 
\paragraph{Performance Comparison to Single-Step Models,} namely \citet{guz-etal-2020-unleashing} and \citet{guz-carenini-2020-coreference}, focusing on the potential of multiple, weak classifiers compared to a single-step, strong classifier. 
\paragraph{Training Time and Size Requirements:} With large models requiring increasingly restrictive training time and resources, the linear combination of weak classifiers allows for more efficient training, making models more accessible, even with severe hardware restrictions. 
\paragraph{Number of Free Parameters:} We plan to investigate the size of weak classifiers in regard to the number of gradient boosting steps and performance. We believe that this detailed investigation can shed further light onto the potential of gradient boosted approaches for discourse parsing. 
\paragraph{Domain Independence Across Steps:} Here, we aim to evaluate whether a larger number of gradient boosting steps leads to increased domain-specificity. This evaluation will compare the model performance of the first $m$ gradient boosting steps (with $1 \leq m \leq n$) to gold-standard discourse structures in different domains, aiming to quantify the correlation of gradient boosted modelling steps with increased domain dependency.

\section{Conclusion}
In this work, we aim to improve current supervised discourse parsers through a gradient-boosted modelling approach and linguistically inspired model adaptions. Compared to previously proposed models, we try to overcome the domain dependency through a staged model capturing increasingly domain-specific information, making better use of the limited amount of gold-standard discourse data. Using more linguistically inspired stack representations and adding a relation classification component, we hope to create a general and domain-independent, fully supervised discourse parser.

\bibliography{anthology,custom}
\bibliographystyle{acl_natbib}

\end{document}